\documentclass[graybox]{svmult}

\usepackage{type1cm}          
\usepackage{makeidx}          
\usepackage{graphicx}         
\usepackage{multicol}         
\usepackage[bottom]{footmisc} 
\usepackage{newtxtext}        %
\usepackage[varvw]{newtxmath} 

\usepackage{marvosym}
\usepackage{glossaries}
\usepackage{makecell}
\usepackage{amsmath} 
\usepackage{amsfonts}
\usepackage{mathrsfs}
\newcommand{\matr}[1]{\mathbf{#1}} 
\newcommand{\vect}[1]{\mathbf{#1}} 
\usepackage{subcaption}

\makeindex             

\begin{document}

\title*{Probing Information Distribution in Transformer Architectures through Entropy Analysis}
\author{Amedeo Buonanno, Alessandro Rivetti, Francesco A. N. Palmieri, \\ Giovanni Di Gennaro, Gianmarco Romano}
\authorrunning{Buonanno et al.}
\institute{
 A. Buonanno (\raisebox{-.15\height}{\large\Letter}) \at Department of Energy Technologies and Renewable Sources, ENEA, Portici (NA), 80055, Italy \email{amedeo.buonanno@enea.it}
 \and A. Rivetti \and F. A. N. Palmieri \and G. Di Gennaro \and G. Romano \at Dipartimento di Ingegneria, Università degli Studi della Campania ``Luigi Vanvitelli'', \\
Aversa (CE), 81031, Italy \\ 
\email{ alessandro.rivetti@studenti.unicampania.it,  \{francesco.palmieri, giovanni.digennaro, gianmarco.romano\}@unicampania.it} 
\and}

\titlerunning{Probing Information Distribution in Transformer Architectures}

\maketitle

\abstract{This work explores entropy analysis as a tool for probing information distribution within Transformer-based architectures. By quantifying token-level uncertainty and examining entropy patterns across different stages of processing, we aim to investigate how information is managed and transformed within these models. As a case study, we apply the methodology to a GPT-based large language model, illustrating its potential to reveal insights into model behavior and internal representations.
This approach may offer insights into model behavior and contribute to the development of interpretability and evaluation frameworks for transformer-based models.
}

\\

\section{Introduction}


Since their introduction, Transformer models \cite{Vaswani2017} have become the cornerstone of modern machine learning, powering a wide range of applications \cite{Islam2024}, including machine translation, time series forecasting \cite{Zhou2021} \cite{Lim2021}, and computer vision \cite{Dosovitskiy2021}. 
Most notably, Transformers serve as the foundational architecture for Large Language Models (LLMs), such as the GPT (Generative Pretrained Transformer) family \cite{Radford2018} \cite{Brown2020} . These models operate by iteratively predicting the next token based on the preceding context and have demonstrated exceptional performance across diverse Natural Language Processing (NLP) tasks \cite{Minaee2025} \cite{Zhao2025}.

Despite their success and ubiquity, the internal mechanisms driving neural networks and Transformer-based models remain largely opaque, raising critical concerns about their transparency and reliability \cite{Huang2025}. 
This has led to a growing body of work aimed at shedding light on its internal behavior. For example many works have been dedicated to unfolding the inner functions of Convolutional Neural Networks \cite{palmieri2019} (and references therein). In studying Transformer architecures, among the most prominent approaches, there are the lens-based methods: techniques that probe intermediate model activations to uncover how predictions evolve during computation.

One such approach is the {\em logit lens} \cite{nostalgebraist2020}, which projects residual stream activations at each layer into the vocabulary space using the model’s output head. 
This reveals a layer-by-layer evolution of the model’s token predictions showing that even early layers exhibit meaningful predictive structure. 
Although insightful, the logit lens has limited precision and provides only a partial view of the underlying computation, it shows "what" the model believes at each stage, but not "how" or "why" those beliefs are updated. 

To address the limitations of logit lens, the tuned lens was introduced as an alternative \cite{Belrose2023}. 
It trains separate affine transformations for each layer of a frozen Transformer, improving the decoding of intermediate representations into token probabilities. 
This yields more accurate reconstructions of latent predictions and reveals structural features aligned with the model’s own computations. 
Additionally, it has been shown to detect adversarial or anomalous inputs based on divergences in prediction trajectories.

A more recent development in this interpretability direction is Entropy-Lens \cite{Ali2025}, a scalable and architecture-agnostic framework that introduces an information-theoretic perspective. 
Entropy-Lens quantifies the Shannon entropy of decoded token distributions at each layer, capturing the evolution of uncertainty throughout the model.
This entropy profile serves as an “information signature” of the Transformer’s computation, correlating with accuracy and revealing differences between model families and task types. 
Crucially, this method requires no model retraining, making it applicable to off-the-shelf LLMs and ViTs alike.

In this work we build upon the recent Entropy-Lens framework.  However, while Entropy-Lens leverages entropy primarily for model or task classification, our focus shifts toward a better understanding of how uncertainty evolves through the network.
Specifically, we study the token-level dynamics of entropy across layers (vertical flow) and token positions (horizontal flow), revealing how the uncertainty evolves within the architecture. 

\section{Problem formulation}
The Transformer model, introduced by Vaswani et al. \cite{Vaswani2017}, represents a paradigm shift in sequence-to-sequence modeling, particularly in NLP. It relies on a mechanism called "attention" to draw global dependencies between input and output. This design allows for significantly more parallelization and has enabled training of larger models on more data, leading to state-of-the-art results across a wide range of tasks \cite{Islam2024}. 
The architecture is typically composed of an encoder and a decoder, each being a stack of identical layers. 

\begin{figure}[!ht]
        \centering
        \includegraphics[width=0.5\linewidth]{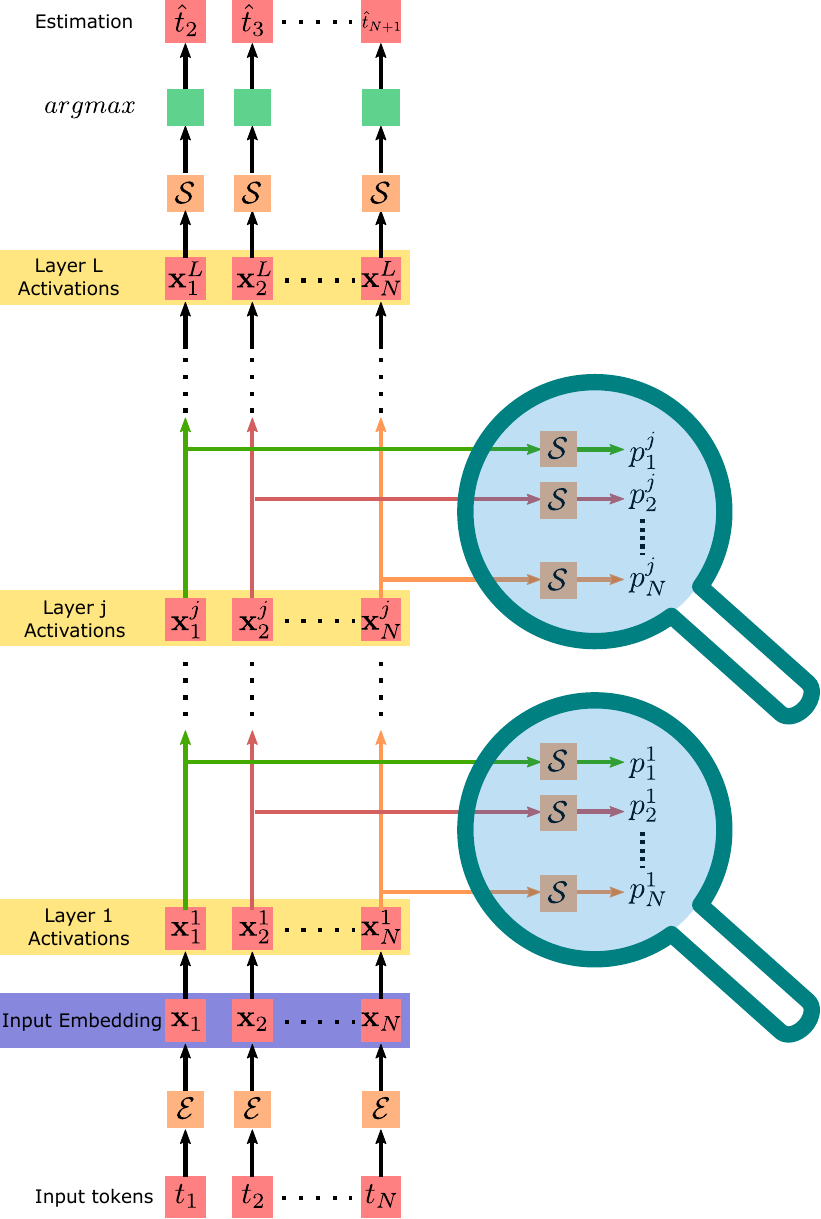}
        \caption{Conceptual representation of a decoder-only Transformer architecture, illustrating the flow of information through multiple processing layers and the lens we employ}
        \label{fig:architecture}
\end{figure}

Each layer in the Transformer stack typically comprises a multi-head self-attention mechanism followed by a position-wise feed-forward network (FFN), along with residual connections and layer normalization \cite{Lin2022}. 
The self-attention module allows the model to weigh the importance of different tokens in the preceding context when forming the representation for the current token, while the FFN provides additional non-linear transformation capacity. 
In this study our focus is on decoder-only architecture whose conceptual representation (regarding the information flow) is in Figure \ref{fig:architecture}.
A key characteristic of the decoder-only Transformers under study, such as GPT-like models, is their autoregressive nature. 
This is typically enforced by a causal attention mask within the self-attention mechanisms, ensuring that the representation of a token at position $i$ is conditioned only on the previous tokens. 
This causal structure is fundamental to their ability to generate coherent text sequentially by predicting one token at a time.

This study aims to leverage entropy as a lens to investigate several key questions: How does token-level uncertainty, as quantified by entropy, evolve across the layers of a decoder-only Transformer model? Are there specific layers that play a more critical role in information consolidation and uncertainty reduction? How does the surrounding contextual information influence this uncertainty profile at different token positions? By addressing these questions, we seek to build a more nuanced understanding of the information flow and a model's internal belief states within these sophisticated architectures.

Let assume to have a discrete vocabulary of tokens, $t \in \mathscr{T}=\{a_1, \ldots, a_M$\}. 
These tokens can be arranged in strings  $(t_1,t_2,\ldots,t_N) \in \mathscr{F}^N \subseteq \mathscr{T}^N$, where $\mathscr{T}^N$ is the set of  all $M^N$ possible sequences.
Each token is  mapped into a $D$-dimensional vector $\vect{x}$ through an operation that we call \emph{embedding}  $\mathcal{E}$ 
\begin{equation}
   \vect{x} = \mathcal{E} (t) \in \mathbb{R}^D  \label{eq:embedding}
\end{equation}
This operation is usually performed via an embedding  matrix $\matr{W}_E  \in \mathbb{R}^{D \times M}$, with $\vect{x} = \matr{W}_E  \vect{t}$, where $ \vect{t}$ is the one-hot vectorial encoding of each token. This corresponds to the memorization of all the token embeddings in the columns of $\matr{W}_E$. This operation can be done sometimes with Word2vec (\cite{Mikolov2013} \cite{digennaro2021a} \cite{digennaro2021b} \cite{digennaro2022a}) and can be chosen apriori according to various criteria, or learned within the whole architecture (as in GPT models).  

The inverse operation, that we call  \emph{unembedding}, returns a distribution of values across $\mathcal{T}$. More specifically, given a generic vector in the embedding space $\vect{y} \in \mathbb{R}^D$, we want to estimate the posterior distribution $p=[P_r(a_1|\vect{y}),...,P_r(a_M|\vect{y})]$ for each token. Such an operation is usually performed using a linear (actually affine) operation followed by a softmax
\begin{equation}
   p = \mathcal{S}(\vect{y}) = S_{max}(\matr{W}_D  \vect{y} + \vect{b}_D), \label{eq:unembedding}
\end{equation}
where $\matr{W}_D \in \mathbb{R}^{M \times D}$ and $\vect{b}_D \in \mathbb{R}^M$ are  a matrix and a bias vector respectively. This is the well-known minimum distance classifier under gaussian likelihood assumption.  Sometimes, as done in GPT-2,  $\matr{W}_D = \matr{W}_E^T$ and $\vect{b}_D =\vect{0}$.

Note that, given the embedding $\vect{x}=\mathcal{E} (t=a_i)$, $p=S(\vect{x})$ does not necessarily return a delta distribution on $a_i$ even if it will mostly concentrated on the $a_i$ token value.
This depends on $M$ and $D$ (i.e. how many tokens are embedded in the $D$-dimensional space and how they are distributed), and what classifier parameters  $\matr{W_D}$ and $\vect{b}_D$ are utilized.   

In any case, this operation, called {\em logit lens}, can be applied to any internal representation in any layer $\{\vect{x}_i^j\}_{i=1,\ldots,N}^{j=1,\ldots,L}$ and any $\{\vect{x}_i\}_{ i=1,\ldots,N}$ of input embeddings, where $L$ is the number of layers and $N$ the number of tokens in the input string. Therefore
\begin{equation}
   p_i^j = \mathcal{S}(\vect{x}_i^j), i=1,\ldots,N, \ j=1,\ldots,L
\end{equation}
can be interpreted as a distribution over the tokens, even if all the transformer operations (residual addition, self-attention, normalization, neural network transformations) are done in the continuous $D$-dimensional space. The logit lens technique is instrumental to our approach, because  probability distributions $p_i^j$  can be interpreted as the model's "current hypothesis", or predictive distribution over the vocabulary, conditioned on the information processed up to that point.

To formalize the variables in the system we denote:
\begin{itemize}
\item  $(T_1,T_2,\ldots,T_Q) \in \mathscr{F}^Q$ a set of $Q$ discrete token random variables;
their values $(t_1,t_2,\ldots,t_Q)$, and $(\overline{t}_1,\overline{t}_2,\ldots,\overline{t}_Q)$ their specific instances. 

\item $(X_1,X_2,\ldots,X_Q) = (\mathcal{E}(T_1),\ldots, \mathcal{E}(T_Q)) $ the set of $Q$ random  $D$-dimensional continuous vectors, which are the embeddings of $(T_1,T_2,\ldots,T_Q)$. 

\item $(X_1^j,X_2^j,\ldots,X_Q^j) $ a set of $Q$ random $D$-dimensional continous vectors in the transformer at layer $j=1,\ldots,L$,

\item  $(T_1^j,T_2^j,\ldots,T_Q^j) = (\mathcal{S}(X_1^j),\ldots, \mathcal{S}(X_Q^j))$ the set of $Q$  discrete token variables at any layer $j=1,\ldots,L$ unembedded from $(X_1^j,X_2^j,\ldots,X_Q^j) $. 
\end{itemize}

Given a specific context string at the input  
$\vect{\overline{t}} = (\overline{t}_1,\overline{t}_2,\ldots,\overline{t}_N)$, 
its embedding $(\vect{\overline{x}}_1,\vect{\overline{x}}_2,\ldots,\vect{\overline{x}}_N)$, 
and all the consequent activations in the networks $\{\vect{\overline{x}}_i^j\}_{i=1,\ldots,N}^{j=1,\ldots,L}$, we can compute the logit distribution related to the input 
($\mathcal{S}(\vect{\overline{x}}_i)=p_i, i=1, \ldots, N$) and at the other layers  
($\mathcal{S}(\vect{\overline{x}}_i^j)=p_i^j(\overline{t}_i,\ldots, \overline{t}_1), i=1,\ldots,N, \ j=1,\ldots,L$) where we have made explicit the causal dependence in the $i$-th stream on the preceding values.
In the last layer we have the estimate:
\begin{equation}
 	\hat{Pr}\{ t_{i+1} | \overline{t}_i,\ldots, \overline{t}_1\} = \mathcal{S}(\vect{\overline{x}}_i^L)=p_i^L(\overline{t}_i,\ldots, \overline{t}_1)
\end{equation}
that is the posterior on token $T_{i+1}$ given the previous ones.
Noteworthy in the last token:
\begin{equation}
	\hat{Pr}\{ t_{N+1}| \overline{t}_{N},\ldots, \overline{t}_1\} = \mathcal{S}(\vect{\overline{x}}_N^L)=p_N^L(\overline{t}_N,\ldots, \overline{t}_1)
\end{equation}
that is the posterior on token $T_{N+1}$ given the previous ones.

To evaluate the spreading of the distributions, we can compute the entropy for each one of them:
\begin{equation}
h_i^j(\overline{t}_i,\ldots, \overline{t}_1) = \sum p_i^j(\overline{t}_i,\ldots, \overline{t}_1) \log_2 \frac{1}{p_i^j(\overline{t}_i,\ldots, \overline{t}_1)}.
\end{equation}
\noindent
Averaging over realizations of strings, following the empirical distribution $p(t_i, \ldots, t_1)$ (related to a specific dataset), we have the average entropy:
\begin{equation}
h_i^j = \mathbb{E}_{p(t_i, \ldots, t_1)}[h_i^j(t_i,\ldots, t_1)].
\end{equation}
\noindent
The information spreading in the intermediate layers shows various degrees of sharpeness in formulating the response in the last layer where:
\begin{itemize}
\item $h_1^L(\overline{t}_1)$: uncertainty in predicting $T_2$ from $\overline{t}_1$ 
\item $h_1^L$: average of the uncertainty in predicting $T_2$ from $T_1$
\item $h_2^L(\overline{t}_2, \overline{t}_1)$: uncertainty in predicting $T_3$ from $\overline{t}_2$ and $\overline{t}_1$
\item $h_2^L$: average of the uncertainty in predicting $T_3$ from $T_2$ and $T_1$
\item $\ldots$
\item $h_N^L(\overline{t}_N, \ldots, \overline{t}_1)$: uncertainty in predicting $T_{N+1}$ from $\overline{t}_N, \overline{t}_{N-1},\ldots, \overline{t}_1$
\item $h_N^L$: average of the uncertainty in predicting $T_{N+1}$ from $T_N, T_{N-1},\ldots, T_1$
\end{itemize}

\section{Results}
This section presents the empirical results obtained from applying the entropy-based analysis presented above to pre-trained GPT-2 models (`small', `medium', `xl')  \cite{Radford2018}. 

We used the High Quality English Sentences dataset \cite{HQESDataset2024}, a curated collection of $1,705,221$ unique sentences originally sourced from C4 \cite{C4Dataset2024} and FineWeb \cite{FineWebDataset2024} corpora.
This dataset was chosen because it ensures a wide diversity of sentences, avoiding biases such as recurring sentence structures or token repetitions, and contains no duplicates. These characteristics make it particularly suitable for our entropy-based analysis.
The dataset is split into $90\%$ train ($1,534,699$ sentences) and $10\%$ test ($170,522$ sentences).
From the train split, we initially considered $30,000$ sentences and filtered them by length, discarding those shorter than $40$ tokens and truncating longer sentences to exactly $40$ tokens. This process resulted in a final set of $4,626$ sentences. 

Our investigation focuses on characterizing the distribution and evolution of information, as quantified by conditional entropy, both across token positions within sequences and through the layers of the network. 

We have used for the logit lens with $\matr{W}_D = \matr{W}_E^T$ and $\vect{b}_D =\vect{0}$.

Our primary analyses are structured around examining the evolution of mean conditional entropy and its full distribution from two main perspectives: across token positions (which we term \emph{horizontal} evolution) and across network layers (\emph{vertical} trajectory).

Firstly, we investigate the horizontal progression by analyzing the average conditional entropy $h_i^j$ as a function of the token position $i$ (i.e., related to the $i$-th token in the input sequence) within each specific layer $j$. 
This allows us to observe, for any given processing depth $j$, how the model's internal uncertainty (as projected onto the vocabulary space) changes as more contextual information from the preceding tokens $\overline{t}_1, \ldots, \overline{t}_i$ is incorporated. 
This is particularly insightful for the final layer ($j=L$), where $h_i^L$ directly relates to the uncertainty in predicting the next token $T_{i+1}$, but the analysis across intermediate layers provides a more comprehensive view of contextual information accumulation throughout the network.

Secondly, we investigate the vertical flow of information by tracking the average conditional entropy $h_i^j$ for specific token positions $i$ as they are processed through the successive layers $j=1, \ldots, L$ of the Transformer (from the initial embedding layer to the final output layer). This "vertical" analysis aims to reveal how uncertainty is modulated and representations are refined at different stages of the network's depth.

Beyond these analyses of mean entropy values, we also explore the full distribution of the conditional entropy $h_i^j(\overline{t}_i, \ldots, \overline{t}_1)$ (i.e., before averaging over sequences).  In this way we are able to illustrate how the distribution of token-level uncertainty changes both horizontally (across different token positions $i$ within a specific layer $j$) and vertically (across different layers $j$ for a specific token position $i$). 
This more granular approach allows for a richer understanding of the model's behavior beyond simple averages, highlighting the consistency or diversity in information processing across different input instances and internal processing stages.
The following subsections will detail the findings from each of these analytical approaches, supported by appropriate visualizations.

\begin{figure}[ht!]
	\centering
	\begin{subfigure}[t]{0.49\textwidth}
    		\includegraphics[width=\linewidth]{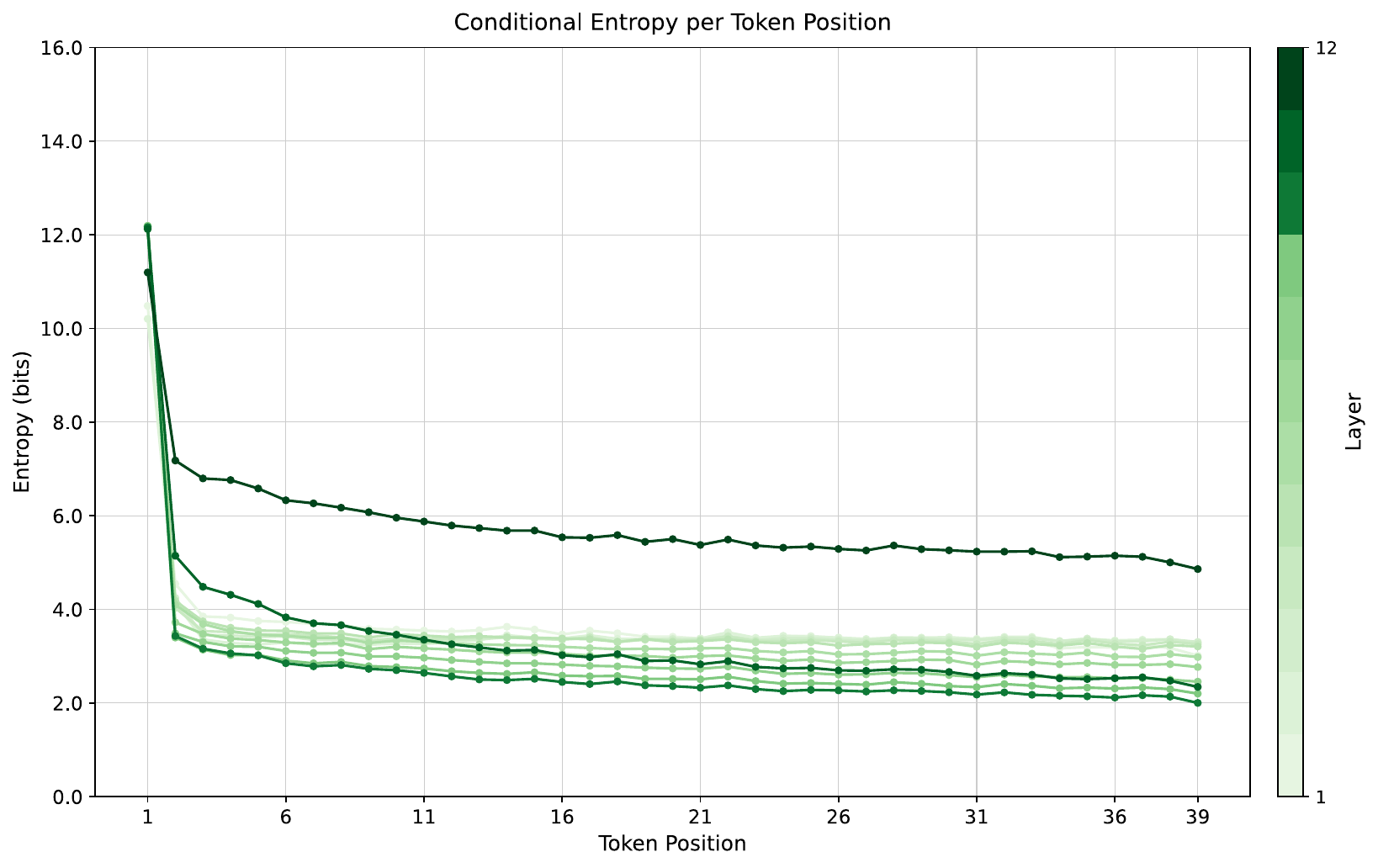}
    		\caption{Average conditional entropy - small}
		\label{fig:entropy_small}
  	\end{subfigure}
  	\hfill
  	\begin{subfigure}[t]{0.49\textwidth}
    		\includegraphics[width=\linewidth]{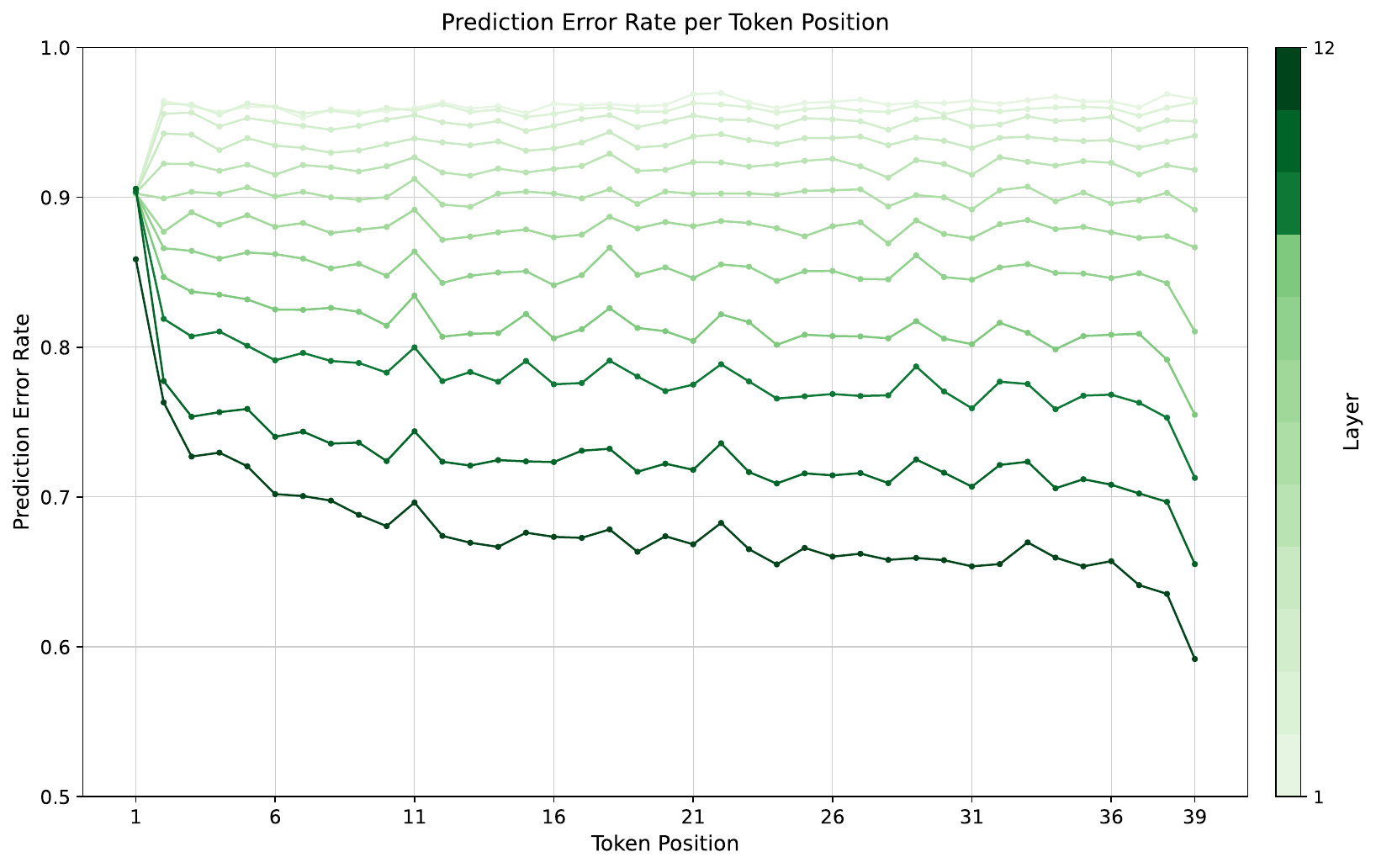}
    		\caption{Prediction errors - small}
		\label{fig:error_small}
  	\end{subfigure}

	 \begin{subfigure}[t]{0.49\textwidth}
    		\includegraphics[width=\linewidth]{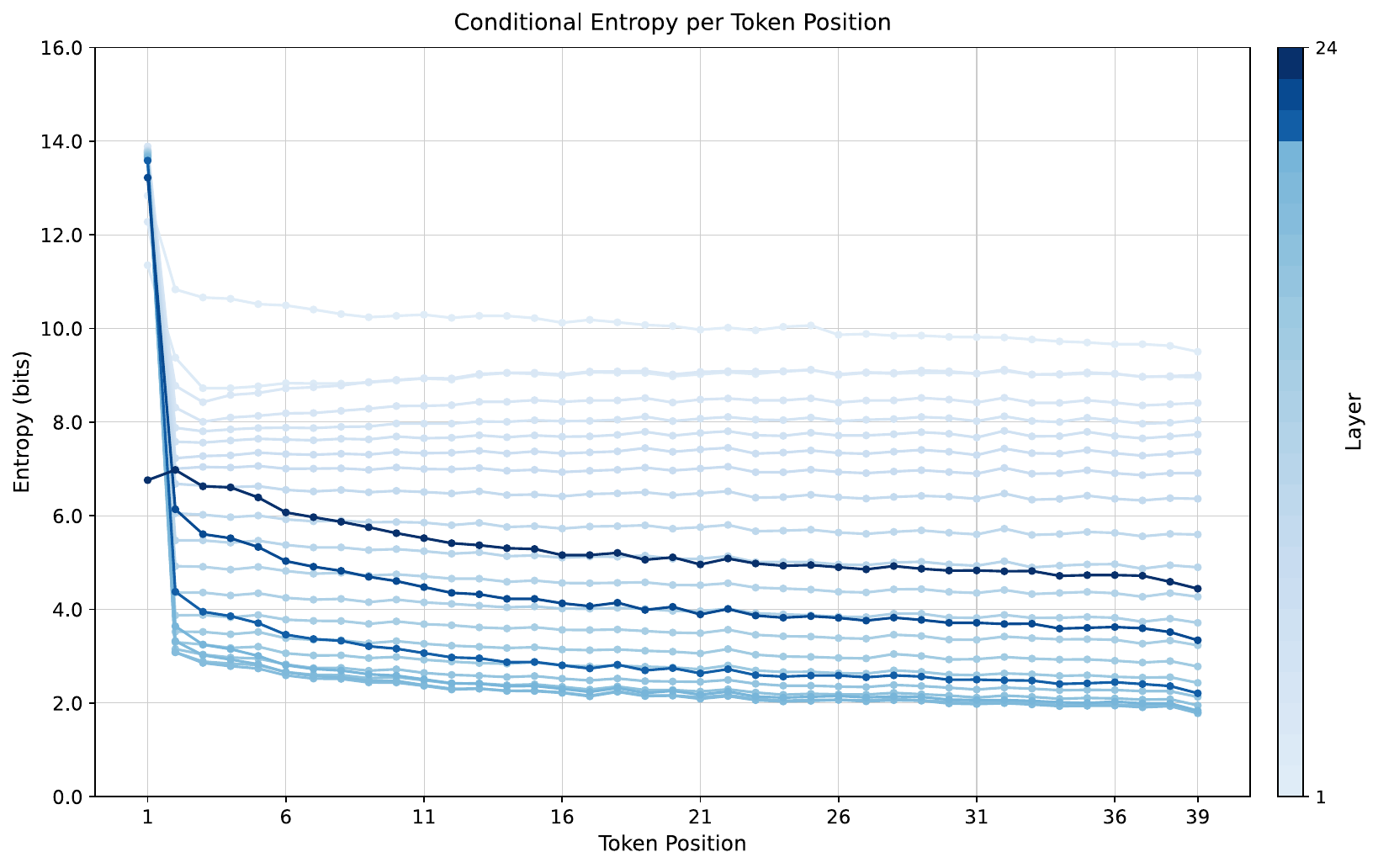}
    		\caption{Average conditional entropy - medium}
		\label{fig:entropy_medium}
  	\end{subfigure}
  	\hfill
  	\begin{subfigure}[t]{0.49\textwidth}
    		\includegraphics[width=\linewidth]{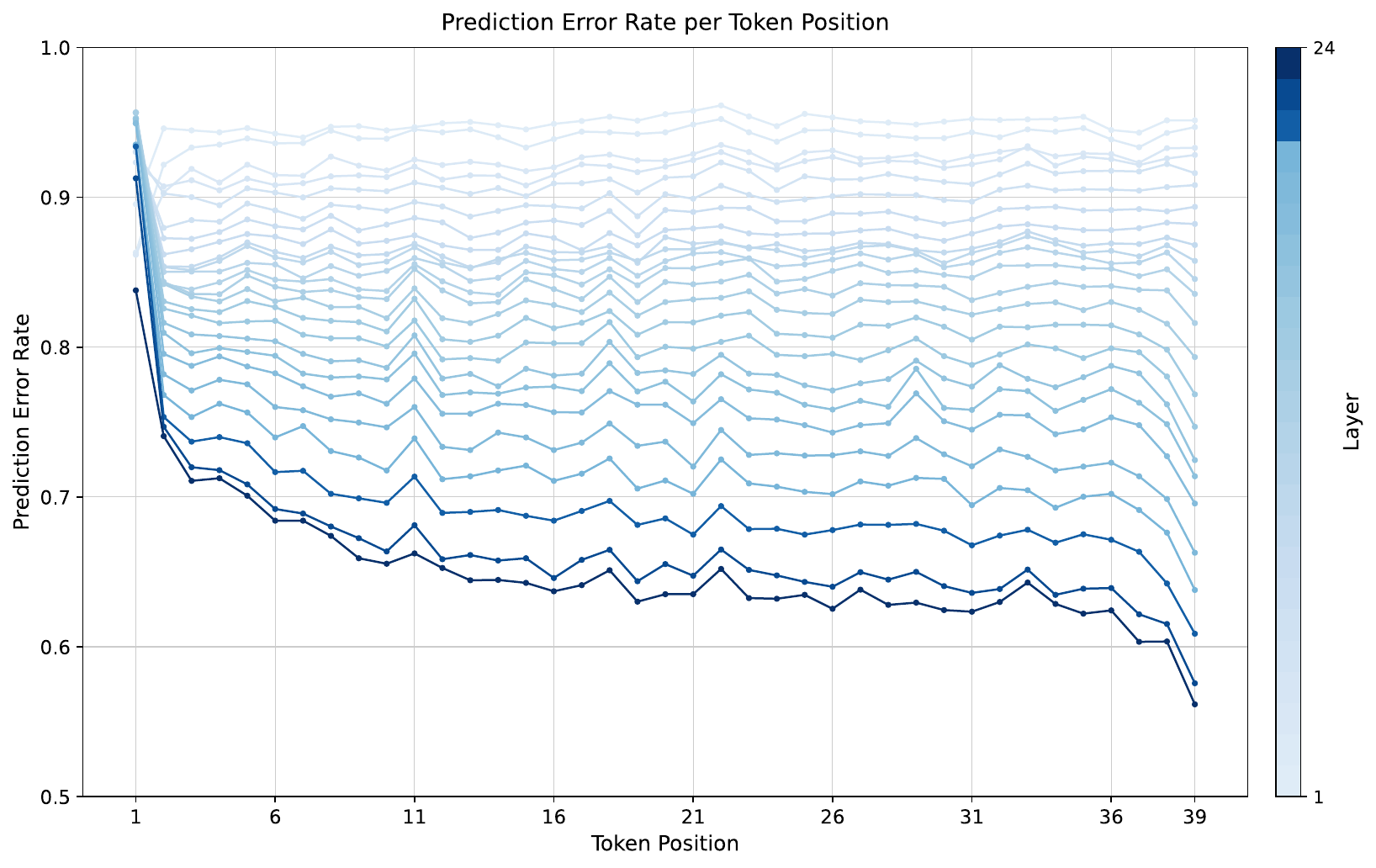}
    		\caption{Prediction errors - medium}
		\label{fig:error_medium}
  	\end{subfigure}

 	\begin{subfigure}[t]{0.49\textwidth}
    		\includegraphics[width=\linewidth]{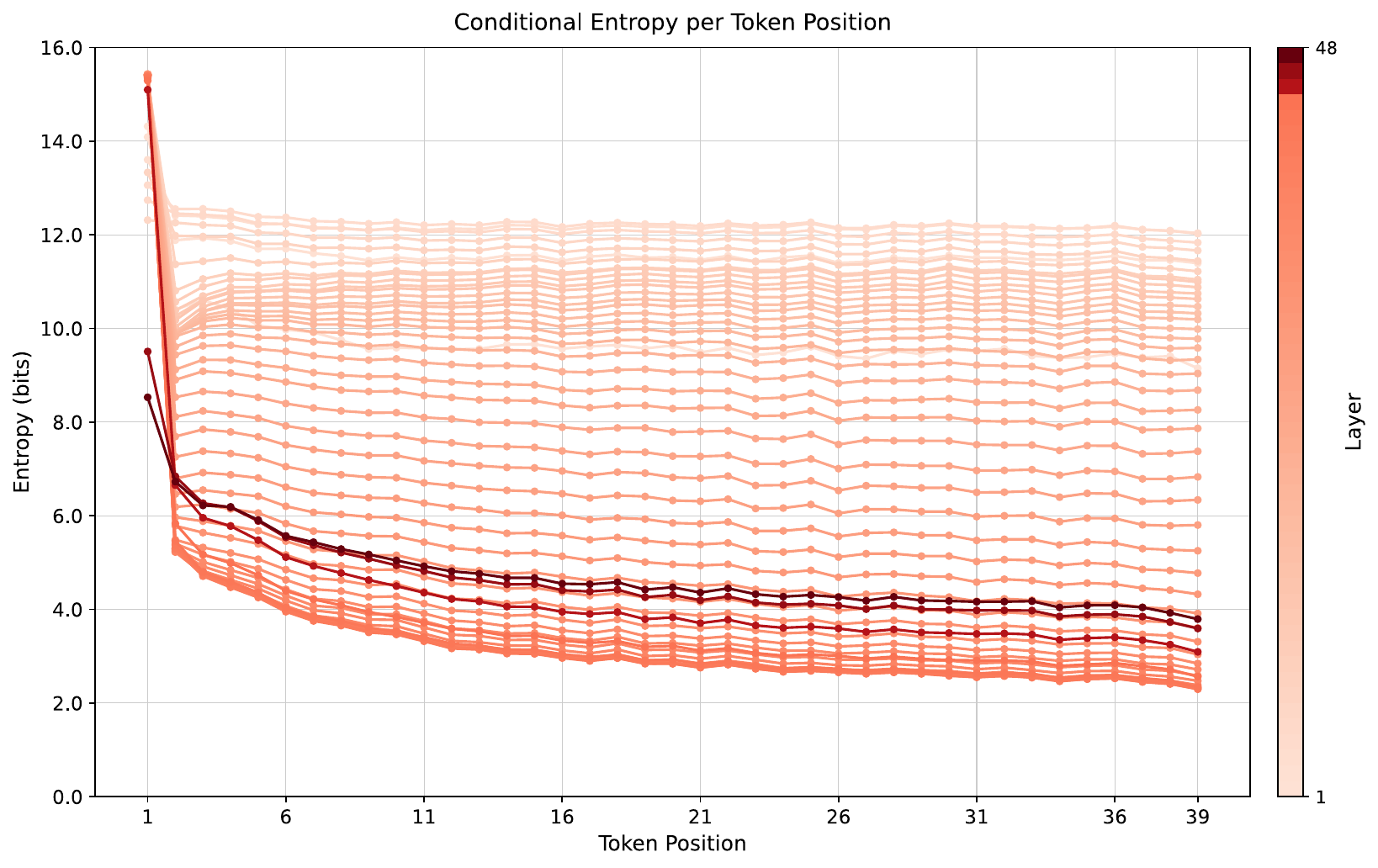}
    		\caption{Average conditional entropy - xl}
		\label{fig:entropy_xl}
  	\end{subfigure}
  	\hfill
  	\begin{subfigure}[t]{0.49\textwidth}
    		\includegraphics[width=\linewidth]{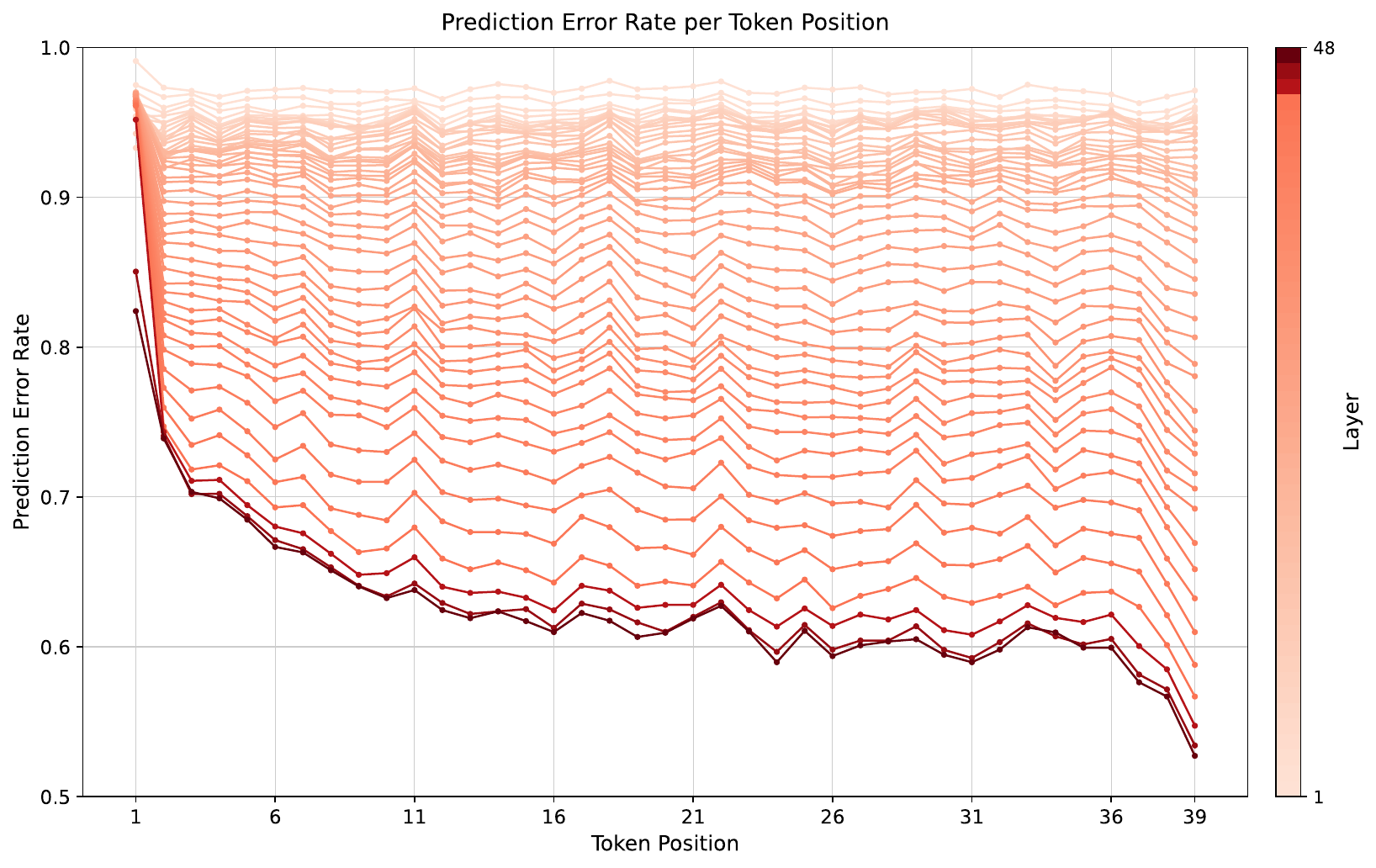}
    		\caption{Prediction errors - xl}
		\label{fig:error_xl}
  	\end{subfigure}
  	\caption{Average conditional entropy and prediction errors at different token positions}
\end{figure}

\begin{figure}[ht!]
	\centering
	\begin{minipage}[b]{0.49\linewidth}
		\vbox to 0.94\textheight{
	    		\includegraphics[width=\linewidth]{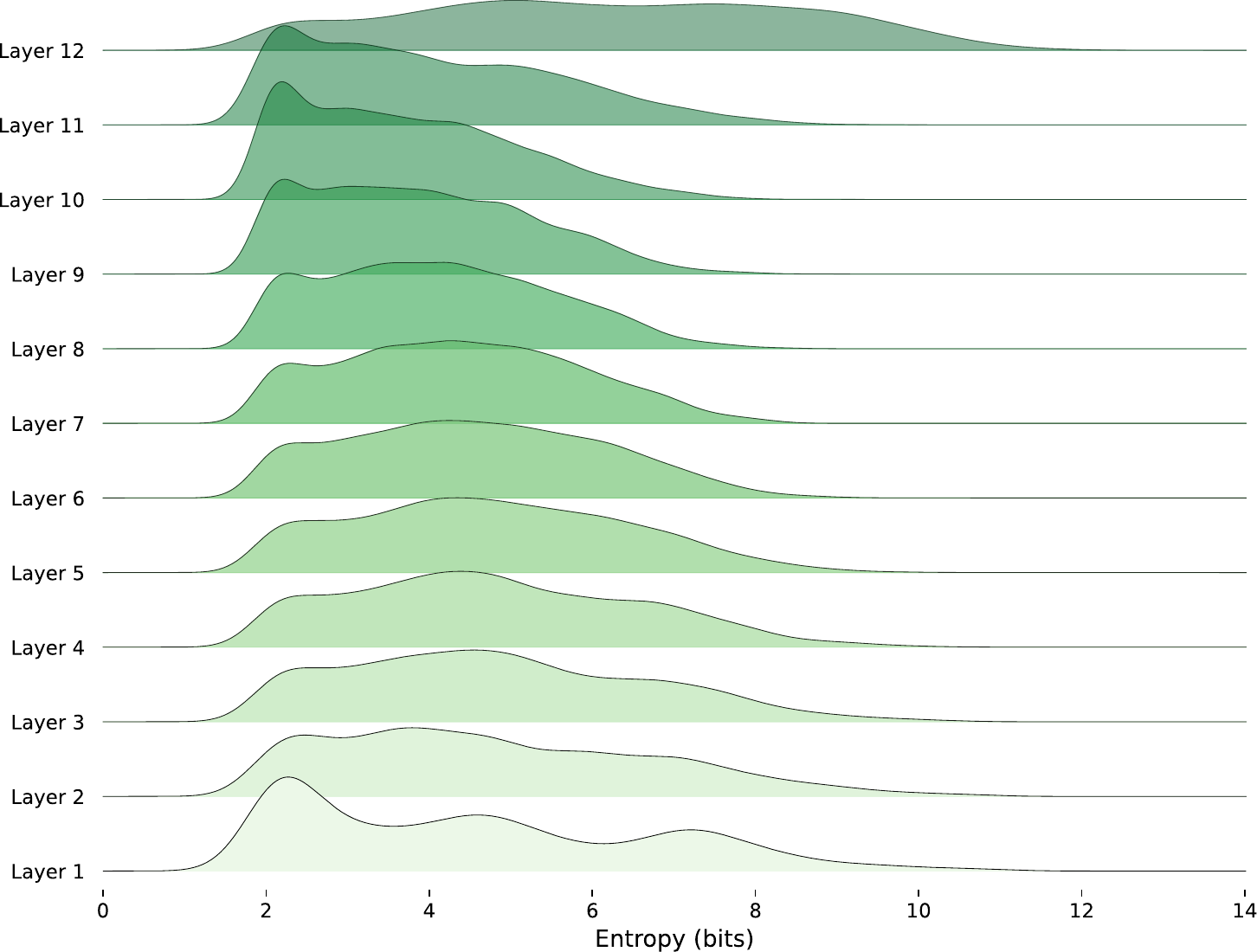}
	    		\caption{Entropy distribution at different layers - small}
	    		\label{fig:ridgeplot_small}
	    		\vfill
	    		\includegraphics[width=\linewidth]{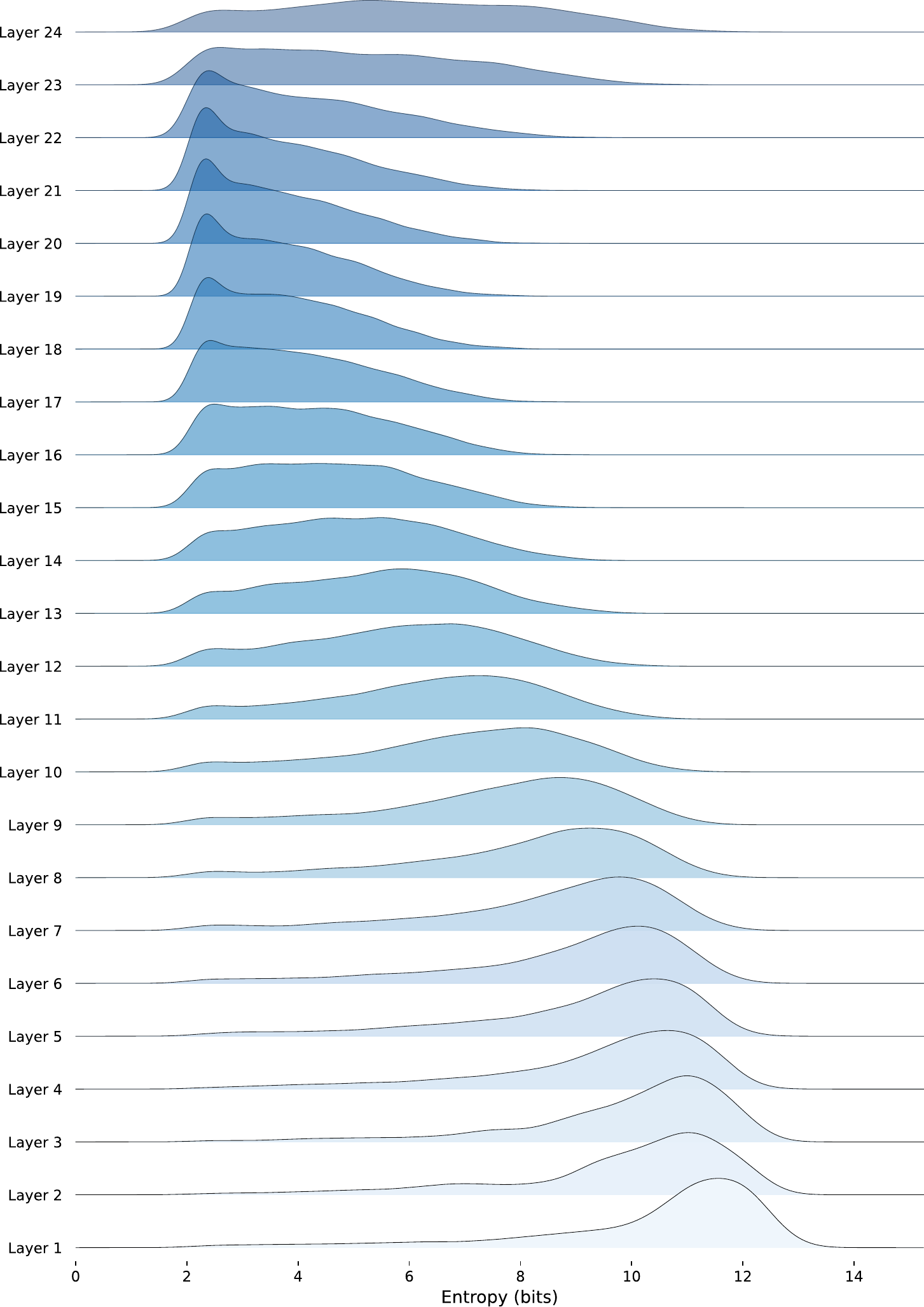}
	    		\caption{Entropy distribution at different layers - medium}
	    		\label{fig:ridgeplot_medium}
		}
	\end{minipage}
	\begin{minipage}[b]{0.49\linewidth}
    		\includegraphics[width=\linewidth]{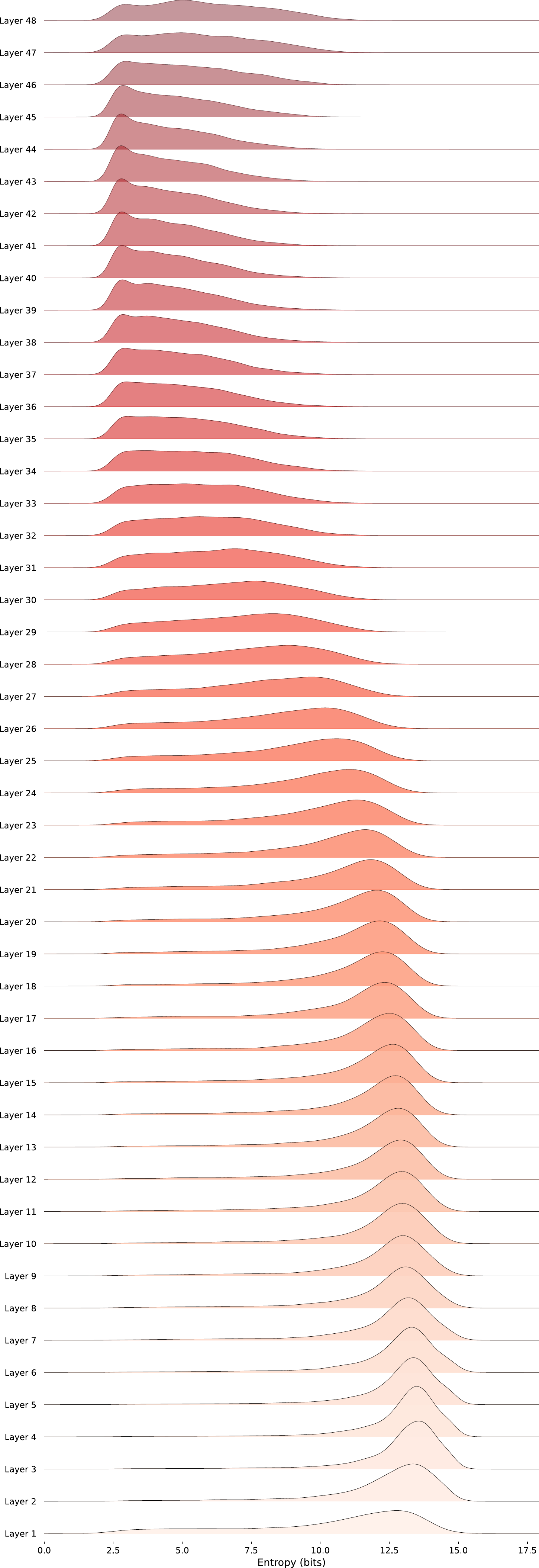}
    		\caption{Entropy distribution at different layers - xl}
    		\label{fig:ridgeplot_xl}
	\end{minipage}
\end{figure}

\section{Discussion}
Across all models (`small': green, `medium': blue, `xl': red), both average conditional entropy (Figure \ref{fig:entropy_small}, \ref{fig:entropy_medium}, \ref{fig:entropy_xl}) and prediction error (Figure \ref{fig:error_small}, \ref{fig:error_medium}, \ref{fig:error_xl})  exhibit a clear decreasing trend along the token position axis: as the context grows, the models become increasingly confident and accurate in predicting the next token. 
This behavior aligns with the autoregressive nature of the models and reflects improved performance with longer input sequences.

When analyzing trends across layers, the prediction error rate decreases monotonically: deeper layers consistently improve top-1 accuracy, suggesting that each layer contributes to progressive refinement of predictions and overall performance.

In contrast, entropy follows a non-monotonic trajectory. 
While early and intermediate layers reduce entropy, indicating growing confidence, the final layers in the `medium' and especially the `xl' models exhibit a slight increase. 
This pattern suggests that the deepest layers might be increasing the model's capacity to consider or assign probability to a broader set of plausible next tokens, potentially encoding semantic alternatives or preparing for more flexible generation, even if the top-1 prediction remains accurate.

This interpretation is further substantiated by the ridgeline plots (Figures \ref{fig:ridgeplot_small}, \ref{fig:ridgeplot_medium}, and \ref{fig:ridgeplot_xl}), which depict the evolution of the conditional entropy distribution for the final token across the layers of the three considered models.

For the `small' model (Figure \ref{fig:ridgeplot_small}), the entropy distributions generally maintain a relatively wide spread and exhibit limited compression throughout the network's depth. Notably, a distinct broadening of the distribution is observed in the very final layer, suggesting an increase in the variability of the model's predictive uncertainty at its ultimate processing stage.

In contrast, the `medium' (Figure \ref{fig:ridgeplot_medium}) and `xl' (Figure \ref{fig:ridgeplot_xl}) models start with markedly high entropy values in their lower layers after which the distribution appears to broaden across several intermediate layers.
This broadening indicates a greater diversity in the levels of uncertainty the model expresses about the next token, depending on the specific input sequence, even as the average predictive entropy concurrently undergoes a gradual decrease.
Moving towards the terminal layers, a phase of distributional compression and mean entropy reduction becomes apparent, signifying that predictions are, on average, becoming more confident.
However, both these larger models exhibit an increase in both the mean and the spread of the predictive entropy distribution in their final one to two layers. 
This terminal-layer phenomenon mirrors the broadening seen in the `small' model, albeit occurring after a more pronounced phase of earlier compression in the larger architectures.

In summary, while prediction accuracy improves consistently with depth, entropy provides a complementary lens into the internal computational dynamics, revealing non-monotonic changes in predictive uncertainty and suggesting that the final layers in larger models might be calibrated for increased representational flexibility or consideration of alternatives.

\section{Conclusion}
In this work, we presented an entropy-based framework to probe the internal information dynamics of Transformer architectures, focusing specifically on decoder-only models such as GPT. By analyzing the evolution of conditional entropy across both token positions and network layers, we gained insight into how uncertainty is managed and transformed throughout the model.
Overall, entropy analysis offers a powerful, architecture-agnostic tool for interpreting the internal states of Transformer models. It complements traditional evaluation metrics and may contribute to developing more interpretable and reliable large language models. 

\begin{acknowledgement}
The work of G. Di Gennaro was supported by the Italian Ministry for University and Research (MUR) - PON Ricerca e Innovazione 2014–2020 (D.M. 1062/2021). 
The work of F.A.N. Palmieri was supported by POR CAMPANIA FESR 2014/2020, A-MOBILITY: Technologies for Autonomous Vehicles.
\end{acknowledgement}

\bibliographystyle{splncs04}
\bibliography{mybibliography}

\begin{thebibliography}{10}
\providecommand{\url}[1]{\texttt{#1}}
\providecommand{\urlprefix}{URL }
\providecommand{\doi}[1]{https://doi.org/#1}

\bibitem{HQESDataset2024}
{agentlans}: high-quality-english-sentences. Hugging Face Dataset (2024),
  available at:
  \url{https://huggingface.co/datasets/agentlans/high-quality-english-sentences}

\bibitem{C4Dataset2024}
{Ai2}: c4. Hugging Face Dataset (2021), available at:
  \url{https://huggingface.co/datasets/allenai/c4}

\bibitem{Ali2025}
Ali, R., Caso, F., Irwin, C., Liò, P.: Entropy-lens: The information signature
  of transformer computations (2025), \url{https://arxiv.org/abs/2502.16570}

\bibitem{Belrose2023}
Belrose, N., Furman, Z., Smith, L., Halawi, D., Ostrovsky, I.V., McKinney, L.,
  Biderman, S., Steinhardt, J.: Eliciting latent predictions from transformers
  with the tuned lens. ArXiv  \textbf{abs/2303.08112} (2023),
  \url{https://api.semanticscholar.org/CorpusID:257504984}

\bibitem{Brown2020}
Brown, T.B., Mann, B., Ryder, N., Subbiah, M., Kaplan, J., Dhariwal, P.,
  Neelakantan, A., Shyam, P., Sastry, G., Askell, A., Agarwal, S.,
  Herbert-Voss, A., Krueger, G., Henighan, T., Child, R., Ramesh, A., Ziegler,
  D.M., Wu, J., Winter, C., Hesse, C., Chen, M., Sigler, E., Litwin, M., Gray,
  S., Chess, B., Clark, J., Berner, C., McCandlish, S., Radford, A., Sutskever,
  I., Amodei, D.: Language models are few-shot learners. In: Proceedings of the
  34th International Conference on Neural Information Processing Systems. NIPS
  '20, Curran Associates Inc., Red Hook, NY, USA (2020)

\bibitem{digennaro2021b}
Di~Gennaro, G., Buonanno, A., Di~Girolamo, A., Ospedale, A., Palmieri, F.A.N.,
  Fedele, G.: An Analysis of Word2Vec for the Italian Language, pp. 137--146.
  Springer Singapore, Singapore (2021). \doi{10.1007/978-981-15-5093-5\_13}

\bibitem{digennaro2021a}
Di~Gennaro, G., Buonanno, A., Palmieri, F.A.N.: Considerations about learning
  word2vec. Journal of Supercomputing  \textbf{77}(11),  12320--12335 (2021).
  \doi{10.1007/s11227-021-03743-2}

\bibitem{digennaro2022a}
Di~Gennaro, G., Ospedale, A., Di~Girolamo, A., Buonanno, A., Palmieri, F.A.,
  Fedele, G.: Split-word architecture in recurrent neural networks pos-tagging.
  In: 2022 International Joint Conference on Neural Networks (IJCNN). pp.
  01--07 (2022). \doi{10.1109/IJCNN55064.2022.9892466}

\bibitem{Dosovitskiy2021}
Dosovitskiy, A., Beyer, L., Kolesnikov, A., Weissenborn, D., Zhai, X.,
  Unterthiner, T., Dehghani, M., Minderer, M., Heigold, G., Gelly, S.,
  Uszkoreit, J., Houlsby, N.: An image is worth 16x16 words: Transformers for
  image recognition at scale. In: 9th International Conference on Learning
  Representations, {ICLR} 2021, Virtual Event, Austria, May 3-7, 2021.
  OpenReview.net (2021)

\bibitem{Huang2025}
Huang, L., Yu, W., Ma, W., Zhong, W., Feng, Z., Wang, H., Chen, Q., Peng, W.,
  Feng, X., Qin, B., Liu, T.: A survey on hallucination in large language
  models: Principles, taxonomy, challenges, and open questions. ACM Trans. Inf.
  Syst.  \textbf{43}(2) (Jan 2025). \doi{10.1145/3703155}

\bibitem{FineWebDataset2024}
{HuggingFace}: Fineweb. Hugging Face Dataset (2024), available at:
  \url{https://huggingface.co/datasets/HuggingFaceFW/fineweb}

\bibitem{Islam2024}
Islam, S., Elmekki, H., Elsebai, A., Bentahar, J., Drawel, N., Rjoub, G.,
  Pedrycz, W.: A comprehensive survey on applications of transformers for deep
  learning tasks. Expert Systems with Applications  \textbf{241},  122666
  (2024). \doi{10.1016/j.eswa.2023.122666}

\bibitem{Lim2021}
Lim, B., Arik, S.O., Loeff, N., Pfister, T.: Temporal fusion transformers for
  interpretable multi-horizon time series forecasting. International Journal of
  Forecasting  \textbf{37}(4),  1748--1764 (2021).
  \doi{10.1016/j.ijforecast.2021.03.012}

\bibitem{Lin2022}
Lin, T., Wang, Y., Liu, X., Qiu, X.: A survey of transformers. AI Open
  \textbf{3},  111--132 (2022). \doi{10.1016/j.aiopen.2022.10.001}

\bibitem{Mikolov2013}
Mikolov, T., Chen, K., Corrado, G., Dean, J.: Efficient estimation of word
  representations in vector space. In: 1st International Conference on Learning
  Representations, {ICLR} 2013, Scottsdale, Arizona, USA, May 2-4, 2013,
  Workshop Track Proceedings (2013), \url{http://arxiv.org/abs/1301.3781}

\bibitem{Minaee2025}
Minaee, S., Mikolov, T., Nikzad, N., Chenaghlu, M., Socher, R., Amatriain, X.,
  Gao, J.: Large language models: A survey (2025),
  \url{https://arxiv.org/abs/2402.06196}

\bibitem{nostalgebraist2020}
nostalgebraist: interpreting gpt: the logit lens.
  \url{https://www.lesswrong.com/posts/
  AcKRB8wDpdaN6v6ru/interpreting-gpt-the-logit-lens}, accessed: 2025-05-25

\bibitem{palmieri2019}
Palmieri, F.A.N., Baldi, M., Buonanno, A., Di~Gennaro, G., Ospedale, F.:
  Probing a Deep Neural Network. Springer Singapore, Singapore (2019).
  \doi{10.1007/978-981-13-8950-4\_19}

\bibitem{Radford2018}
Radford, A., Narasimhan, K., Salimans, T., Sutskever, I.: Improving language
  understanding by generative pre-training  (2018)

\bibitem{Vaswani2017}
Vaswani, A., Shazeer, N., Parmar, N., Uszkoreit, J., Jones, L., Gomez, A.N.,
  Kaiser, L.u., Polosukhin, I.: Attention is all you need. In: Guyon, I.,
  Luxburg, U.V., Bengio, S., Wallach, H., Fergus, R., Vishwanathan, S.,
  Garnett, R. (eds.) Advances in Neural Information Processing Systems.
  vol.~30. Curran Associates, Inc. (2017)

\bibitem{Zhao2025}
Zhao, W.X., Zhou, K., Li, J., Tang, T., Wang, X., Hou, Y., Min, Y., Zhang, B.,
  Zhang, J., Dong, Z., Du, Y., Yang, C., Chen, Y., Chen, Z., Jiang, J., Ren,
  R., Li, Y., Tang, X., Liu, Z., Liu, P., Nie, J.Y., Wen, J.R.: A survey of
  large language models (2025), \url{https://arxiv.org/abs/2303.18223}

\bibitem{Zhou2021}
Zhou, H., Zhang, S., Peng, J., Zhang, S., Li, J., Xiong, H., Zhang, W.:
  Informer: Beyond efficient transformer for long sequence time-series
  forecasting. In: Thirty-Fifth {AAAI} Conference on Artificial Intelligence,
  {AAAI} 2021, Virtual Event, February 2-9, 2021. pp. 11106--11115. {AAAI}
  Press (2021). \doi{10.1609/AAAI.V35I12.17325}

\end{thebibliography}

\end{document}